\documentclass{article}
\usepackage{ijcai24}

\usepackage{times}
\usepackage{soul}
\usepackage{url}
\usepackage[hidelinks]{hyperref}
\usepackage[utf8]{inputenc}
\usepackage[small]{caption}
\usepackage{graphicx}
\usepackage{amsmath}

\DeclareMathOperator*{\argmax}{arg\,max}
\DeclareMathOperator*{\argmin}{arg\,min}

\newcommand{\fr}{\mathbin{\mathbf{\mathbf{fr}}}}
\newcommand{\mr}{\mathbin{\mathbf{\mathbf{mr}}}}
\usepackage{amssymb}
\usepackage{amsthm}
\usepackage{booktabs}
\usepackage{algorithm}
\usepackage[noend]{algorithmic}
\algsetup{linenosize=\footnotesize}

\usepackage{multirow}

\usepackage[switch]{lineno}
\usepackage{tikz}
\input{tikz_config}
\usepackage{xspace}
\usepackage{todonotes}
\usepackage{makecell}
\newcommand{\lime}{\textsc{Lime}\xspace}
\newcommand{\states}{\mathcal{S}}
\newcommand{\actions}{\mathcal{A}}
\newcommand{\trans}{\mathcal{P}}
\newcommand{\reward}{\mathcal{R}}
\newcommand{\mdp}{\mathcal{M}}

\newcommand{\dist}{\ensuremath{\Delta}}

\newcommand{\rlrule}{\rho}

\usepackage{amsthm}
\newtheorem{definition}{Definition}
\usepackage{wrapfig}

\newcommand{\algname}{\mbox{\textsc{Legible}}}

\title{Rule-Guided Reinforcement Learning Policy Evaluation and Improvement}
\author{
Martin Tappler$^1$
\and
Ignacio D. Lopez-Miguel$^1$\and
Sebastian Tschiatschek$^{2}$\And
Ezio Bartocci$^1$\\
\affiliations
$^1$TU Wien\\
$^2$University of Vienna
\emails
\{martin.tappler, ignacio.lopez, ezio.bartocci\}@tuwien.ac.at,
sebastian.tschiatschek@univie.ac.at
}

\begin{document}

\maketitle

\begin{abstract}

% This paper proposes an approach to utilizing domain knowledge to evaluate and improve RL policies. 
We consider the challenging problem of using domain knowledge to improve deep reinforcement learning policies. To this end, we propose \algname{}, a novel approach, following a multi-step process, which starts by mining rules from a deep RL policy, constituting a partially symbolic representation.  These rules describe which decisions the RL policy makes and which it avoids making. In the second step, we generalize the mined rules using domain knowledge expressed as metamorphic relations. We adapt these relations from software testing to RL to specify expected changes of actions in response to changes in observations. The third step is evaluating generalized rules to determine which generalizations improve performance when enforced. These improvements show weaknesses in the policy, where it has not learned the general rules and thus can be improved by rule guidance. \algname{} supported by metamorphic relations provides a principled way of expressing and enforcing domain knowledge about RL environments. We show the efficacy of our approach by demonstrating that it effectively finds weaknesses, accompanied by explanations of these weaknesses, in eleven RL environments and by showcasing that guiding policy execution with rules improves performance w.r.t. gained reward. 
%We also exemplarily highlight the potential of the type of mined rules for explaining RL policies. 
\end{abstract}

\section{Introduction}
Deep reinforcement learning (RL) has shown impressive feats. Starting from deep Q-learning~\cite{DBLP:journals/nature/MnihKSRVBGRFOPB15}, where a single network architecture successfully played dozens of Atari games, deep RL has moved on to more complex domains and achieved super-human performance in games, like Go~\cite{DBLP:journals/nature/SilverHMGSDSAPL16} and chess~\cite{doi:10.1126/science.aar6404}, and it mastered StarCraft~\cite{DBLP:journals/nature/VinyalsBCMDCCPE19}. Despite the impressive success in winning strategy games, deploying deep RL agents faces challenges in other domains. First, the decision-making of RL agents needs to be more accurate, making it hard to trust their decisions. Second, successful applications are often not the result of RL alone. 

A thorough analysis of an agent's behavior may help understand its "reasoning" -- e.g., chess masters found that AlphaZero highly values king mobility~\cite{alpha_zero_news}. Explainable AI (XAI) methods strive to alleviate the analysis of policies by providing human-understandable explanations of an RL agent's decision. 
Relating to the second point, AlphaZero combines RL with Monte Carlo tree search (MCTS) to better evaluate situations. Thus, it considers the application domain, as MCTS works well for decision-making in board games~\cite{DBLP:conf/aiide/ChaslotBSS08}. 
% The former requires analysis by domain experts, while the latter takes the application domain into account, as MCTS is known to work well for decision-making in board games.\todo{do we need a reference?}  

\noindent
\textbf{Contributions.}
We propose \textbf{poLicy Evaluation GuIded By ruLEs (\algname{})}, a rule-based framework to create explanations of an RL policy's decisions and to evaluate and improve the RL policy under consideration, supported by domain knowledge.
% integrate domain knowledge into the policy's decision-making. 
\algname{} comprises three main methods that build upon each other. \textbf{1. Mining Rules:} We first mine rules that partially represent the policy. These rules are split into positive, denoting action choices, and negative rules, denoting action avoidance in particular situations. They enable human comprehension of a policy's decisions and symbolic reasoning and manipulation. \textbf{2. Generalizing Rules:} We propose an approach to generalize mined rules to other situations through domain knowledge about symmetries and relations known about the environments. Borrowing the concept from software testing, we formalize the generalization via metamorphic relations~\cite{DBLP:journals/csur/ChenKLPTTZ18}. \textbf{3. Rule-Guided Execution: } Finally, we propose to guide the execution of a policy in the RL environment with generalized rules, i.e., enforce the decisions prescribed by rules, which serves two purposes. \textbf{3.1. Evaluation: } Rule-guided execution helps to identify weaknesses in the policy. If we find that enforcing generalized rules corresponding to a certain rule $r$ improves the RL agent's performance, we can deduce that the original rule $r$ is likely adequate. However, the policy has not learned to decide adequately in related situations, i.e., we identify a weakness in the policy's generalization. Additionally, the generalized rules explain the cause of the weakness. \textbf{3.2. Policy Improvement: } Finally, guiding policy execution with the composition of generalized rules, which the evaluation deemed useful, creates a new policy that improves upon the original policy. We demonstrate these aspects of \algname{} in experiments with policies trained in six PAC-Man RL environments~\cite{berkeley_pacman} and five environments from Farama's \texttt{highway-env}~\cite{highway_env}. 

% \todo{Paper structure?}

% Through the execution of a trained deep RL policy guided by generalized rules, we identify weaknesses in the policy. Weaknesses are situations where enforcing a generalized rule achieves better performance than following the decisions of the policy. Hence, these weaknesses show that the policy learned one aspect of its environment well, but not a related aspect. 
% The rules essentially serve two purposes: (1) they explain what the policy has or has not learned -- in cases. where execution guided by a generalized rule yields better performance -- and (2) they provide a mechanism to improve the deep RL policy. Training guided by rules can help to integrate those rules into the policy.  

\paragraph{Related Work.}
Our work is related to explainable RL (XRL), runtime enforcement, and evaluation of RL policies. \cite{DBLP:journals/csur/MilaniTVF24} provide an excellent survey on XRL including a taxonomy and evaluation criteria. Learning rules falls into the most popular taxonomic category of \emph{feature importance} and the subcategory \emph{Convert Policy to an Interpretable Format}. Although symbolic rules as an interpretable format are not popular yet,  decision trees which are related to rules are often used~\cite{DBLP:conf/aaai/BewleyL21,9925786,DBLP:conf/nips/BastaniPS18,DBLP:conf/pkdd/MilaniZTSKPF22}. Several neuro-symbolic approaches have been proposed for RL, where neural networks encode symbolic relations and logical rules. Examples include neural logic machines~\cite{DBLP:conf/iclr/DongMLWLZ19} and neural logic RL~\cite{DBLP:conf/icml/JiangL19}, and relational approaches~\cite{DBLP:conf/nips/DelfosseSDK23,DBLP:journals/corr/abs-1806-01830} that enable the extraction of symbolic information and rules from learned policies. 
In contrast to these works, we consider standard architectures used in deep RL and extract partial rule-based representations of learned policies. Moreover, we demonstrate the application of rules beyond explanations for evaluation and runtime enforcement. 

There are two main strands of work in RL policy evaluation: off-policy evaluation (OPE) and testing of RL policies. OPE~\cite{DBLP:journals/corr/abs-2212-06355,DBLP:conf/nips/ChandakNSLBT21,DBLP:conf/icml/JiangL16} estimates the expected performance of a new policy using existing data from a previously learned policy. Since we generate new data, our approach to evaluation is closer to RL testing~\cite{DBLP:conf/ijcai/TapplerCAK22,DBLP:conf/icst/TapplerMAK24,DBLP:journals/tse/ZolfagharianABBS23,DBLP:conf/kbse/LiWZCCZXMZ23,DBLP:journals/tosem/BiagiolaT24}, which creates challenging situations to test policies in them. These approaches apply software testing concepts, like search-based testing, to RL. To our knowledge, we are the first to apply metamorphic testing in RL, which is popular for testing other machine-learning models~\cite{DBLP:journals/tse/ZhangHML22,DBLP:journals/jss/XieHMKXC11,DBLP:conf/kbse/GuoXLZLLS20} due to its applicability when absolute correctness criteria are not available. To test policies, we perform rule-guided execution of policies, which can be considered a type of runtime enforcement~\cite{DBLP:conf/rv/Falcone10}. In RL, runtime enforcement has gained popularity in the form of \emph{shielding}~\cite{DBLP:conf/aaai/AlshiekhBEKNT18,DBLP:conf/sii/OdriozolaOlaldeZA23}, which enforces pre-specified properties like safety. In contrast, we enforce generalizations of learned rules to evaluate these generalizations. 
\section{Preliminaries}
\subsection{Reinforcement Learning}
% In RL, an agent learns a decision-making policy for a given task by trial and error. At each step, the agent observes the environment's state and performs an action, which triggers a stochastic state transition. It then receives feedback in the form of a numerical value, called reward, which tells the agent how well it is doing, and the new state of the environment. During training, the agent learns how to maximize the cumulative reward it gets.

An RL agent learns a decision-making policy for a task by trial and error. At each step, the agent observes the environment's state and performs an action, triggering a stochastic state transition. It then receives feedback in the form of a numerical value, called reward, telling it how well it is doing, and the new state of the environment. During training, the agent learns how to maximize the cumulative reward it gets.

Let $\dist(S)$ denote a probability distribution over a set $S$. Formally, an agent interacts with a Markov decision process (MDP) $\mdp = \langle \states, s_0, \actions, \trans, \reward \rangle$ consisting of a set of states $\states$, an initial state $s_0 \in \states$, a set of actions $\actions$, a probabilistic transition function $\trans:\states \times \actions \rightarrow \dist(\states)$, and a reward function $\reward: \states \times \actions \times \states \rightarrow \dist(\mathbb{R})$. The agent's behavior is characterized by a policy $\pi: \states \rightarrow \dist(\actions)$, defining a distribution over actions to take in given states. Executing a policy $\pi$ within an environment modeled by an MDP $\mdp$ 
%induces a Markov chain $\mdp^\pi$ and 
yields traces of the form $s_0, a_0, r_1, s_1, \ldots, a_{n-1}, r_n, s_n$ with $r_i \sim \reward(s_{i-1}, a_{i-1}, s_i)$, $\trans(s_i,a_i)(s_{i+1}) > 0$, and $\pi(s_i)(a_i) > 0$. Such a finite execution is also called an episode. The agent's goal is to learn a policy that maximizes the discounted cumulative reward $R =\mathbb{E}(\sum_{i=0}^\infty \gamma^i r_{i+1})$ for a discount factor $\gamma \in [0,1]$. 
% Additionally, we consider a special set of terminal states $\mathcal{S}_U \subset \states$ in this paper. These are states that terminate an episode when the agent reaches them, causing it to fail its task. We call them unsafe states.

To handle large state spaces, RL algorithms often employ neural networks as function approximators, e.g., for state-action value functions~\cite{DBLP:journals/nature/MnihKSRVBGRFOPB15}. Since we mostly focus on extracting symbolic knowledge from policies in this paper, we do not detail the specifics of RL algorithms. 

\noindent
\textbf{State-Action Value Function.}
Q-learning~\cite{DBLP:journals/ml/WatkinsD92} and its deep variants~\cite{DBLP:journals/nature/MnihKSRVBGRFOPB15,DBLP:conf/aaai/HasseltGS16} are based on the notion of Q-function, or state-action value function. This function is defined as $Q_\pi(s,a) = \mathbb{E}[\sum_{i=0}^\infty \gamma^i r_{i+1}| s_0 = s, a_0 = a]$, i.e., it is the expected return after executing $a$ in state $s$ while following policy $\pi$. We focus mainly on Q-learning agents and use the Q-function to identify actions that an agent avoids in a given situation, i.e., actions with low Q-values. 

\subsection{Explainable \& Interpretable AI}
Explainable AI (XAI) methods create human-understandable explanations of individual decisions or predictions of models that are otherwise not interpretable, like neural networks, while interpretability techniques commonly extract useful information, like interpretable surrogate models, from non-interpretable models~\cite{molnar2022}.
To enable human comprehension, explanations commonly focus on the most relevant factors leading to a decision, like the most important input features. This working principle makes XAI techniques good candidates for creating abstractions for symbolic AI. We use a model-agnostic XAI technique for explaining individual predictions, called \lime~\cite{DBLP:conf/kdd/Ribeiro0G16}, and we use rules as (partial) surrogate models of RL policies.

\noindent
\textbf{\lime.}
% Consider that we want to explain why a policy $\pi$ has chosen action $a$ in state $s$. Treating this as a classification task, 
Treating the choice of action $a$ through a policy $\pi$ in state $s$ as a classification task, \lime can learn a local interpretable surrogate model around $s$. It samples new data points around $s$ and queries $\pi$ to learn the surrogate model, which explains what features of $s$ lead to the choice of $a$ and provides weights describing the strength of influence. As we consider states that are vectors containing information about the environment, we use \lime for tabular data.  In this case, it samples new values in the neighborhood of numerical features and values from a \emph{training set} for categorical features. 

We use the Python implementation of \lime~\cite{python_lime}, which provides an explanation for predicting every available class, i.e., for choosing every available action. 
% We denote calls to \lime by $\textit{exp} \gets \lime(s,\pi,n_\mathit{fe})$, where $s$ is the state for which we want explanations, $\pi$ is the policy under consideration, and $n_\mathit{fe}$ is the number of features we want in the explanation $\mathit{exp}$. Additionally, we use $\lime\mathit{Intervals}(j)$ to denote the intervals computed by \lime to discretize the numerical feature $j$.
Several alternatives to \lime exist, like SHAP~\cite{DBLP:conf/nips/LundbergL17}, but we have chosen \lime as it is reasonably fast and showed promising results in experiments. Since it is model-agnostic and works for image and textual data, \lime enables us to easily adapt to changing representations of $\pi$ and to handle additional types of features. 

\noindent
\textbf{Interpretability \& Rules.}
% Interpretability is approached in different ways in RL, for example, by learning policies that are inherently interpretable, like decision trees~\cite{DBLP:journals/csur/MilaniTVF24}. Alternatively, post-hoc interpretability for a DRL $\pi$ may be achieved by creating a decision tree policy from $\pi$ through imitation learning~\cite{DBLP:journals/csur/MilaniTVF24,viper_in_milani}.
RL interpretability is approached in different ways, e.g., post-hoc interpretability for a DRL policy $\pi$ may be achieved by learning a decision tree policy from $\pi$ through imitation learning~\cite{DBLP:conf/nips/BastaniPS18}. We learn rule-based representations of policies, which, like decision trees, enable manual comprehension and symbolic reasoning~\cite{DBLP:journals/fgcs/ApteW97}. The rules are of the form $a \leftarrow c_1, \ldots, c_n$, specifying to take action $a$ if conditions $c_1$ to $c_n$ hold. We have chosen rules, as they conveniently enable learning partial representation of a policy. For rule learning, we use the algorithm RIPPER~\cite{DBLP:conf/icml/Cohen95}, where we treat the selection of actions as a classification task from states to actions. To clearly distinguish between reinforcement learning and rule learning, we refer to the latter as rule mining.

\subsection{Metamorphic Testing}
Metamorphic testing (MT)~\cite{DBLP:journals/csur/ChenKLPTTZ18} is a technique for generating test cases and deciding on test verdicts. It is based on metamorphic relations (MRs), which define a relation between a sequence of a program's inputs and the corresponding outputs. MRs express the output changes in response to input changes. Consider, e.g., a program implementing the factorial, an MR could be defined as $\langle (n,n+1), (r,r\cdot(n+1))\rangle$, where the first pair represents two inputs and the second pair represents two outputs denoting that if $n! = r$ then $(n+1)! = r\cdot (n+1)$ should hold. Note that MRs do not specify correctness criteria in absolute terms but as relations. This makes MT a popular choice for deciding on test verdicts in machine learning~\cite{DBLP:journals/tse/ZhangHML22}, e.g., in image recognition~\cite{DBLP:conf/issta/DwarakanathASRB18}. While it is hardly possible to completely characterize a cat, MRs can express properties like \emph{if an image shows a cat, then a rotated version of that image still shows a cat}. 
In this paper, we adapt MRs from programs to RL agents, by forming relations over states (inputs to the agent) and actions (outputs of the agent) to express domain knowledge. Unlike in MT, we do not make assumptions about the correctness of specific chosen actions. % We use a secondary criterion to determine if an MR holds.

\begin{figure*}[t]
    \centering
\input{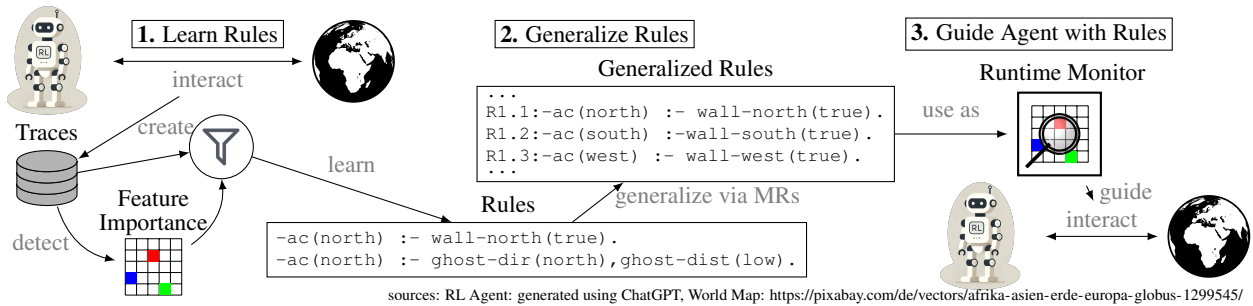} \vspace{-0.3cm}
    \caption{Overview of the proposed approach}
    \label{fig:overview}
\end{figure*}

\section{\algname{}}
This section presents an overview of \textbf{poLicy Evaluation GuIded By ruLEs (\algname{})}, our rule-based framework for policy evaluation and policy improvement, which comprises three main steps, depicted in Fig.~\ref{fig:overview}. Given a deep RL policy, Step 1 learns symbolic rules that (partially) capture the decision-making of the policy. Rules are horn clauses where the head specifies an action and the body defines states where the rule should be applied. We distinguish between positive rules, for situations where the policy chooses a certain action, and negative rules describing situations where it avoids a certain action. Rule mining applies two types of criteria: (1) the rules should reflect what features are important to the decisions of the RL policy under consideration and (2) the rules should be accurate and cover as many situations as possible, where we favor accuracy. Hence, mined rules explain which decisions are important and occur often. 

Step 2 generalizes rules to new situations that are related to the originally covered situations through user-specified relations.  For this purpose, we use metamorphic relations (MRs)~\cite{DBLP:journals/csur/ChenKLPTTZ18}
from software testing, which usually specify how program outputs should change in response to input changes. Likewise, we use them to specify how decisions should change in response to a change in the state. MRs generally need to be created manually and they reflect some domain knowledge about the environment, like symmetry constraints, i.e., through Step 2, we provide a method to introduce symbolic domain knowledge into RL. 

Finally, in Step 3 we apply the generalized rules during executions of the RL policy under consideration. Actions prescribed by rules generalized from positive rules are enforced while actions of negative generalized rules are blocked. In this way, we determine which rules generalize to improved behavior in the RL environment. Conversely, such improvements reveal weaknesses in the RL policy since they show that the policy makes useful decisions in some specific situations but it does not generalize these decisions to all related situations. Hence, Step 3 provides a way to evaluate RL policies and a basis for policy improvements by enforcing sets of generalized rules that influence behavior positively. 

% The first two steps of this process create rules that explain the behavior of RL agents and how it may be generalized. The third step helps identify potential weaknesses of the RL agent. that may be mitigated through extended training with generalized rules being applied. 

\noindent
\textbf{Setting.}
For the remainder of this paper, let $\pi$ be the policy of a Q-learning-based agent 
%trained in an environment $\mathcal{M} =\langle S,\iota, A, \mathcal{P}, R\rangle$, 
and $Q\colon S \times A \rightarrow \mathbb{R}$ be its Q-function. A state $s$ is a vector in $\mathbb{R}^n$, where $n$ is the number of features, $f_i$ for $i \in [1..n]$  denotes the $i$\textsuperscript{th} feature, and $f_i^s$ denotes its value in $s$.
To enable rule learning, we consider environments with a discrete action space and discretize the state space by defining intervals for every feature separately. To simplify the presentation, we write $f_i^s = k$ to denote $f_i^s \in [l_{i,k},u_{i,k}]$ where $[l_{i,k},u_{i,k}]$ is the $k$\textsuperscript{th} interval for feature~$i$.

\noindent
\textbf{Running Example.}
\begin{figure}
    \centering
    \includegraphics[width=.3\textwidth,trim=12 0 10 20,clip]{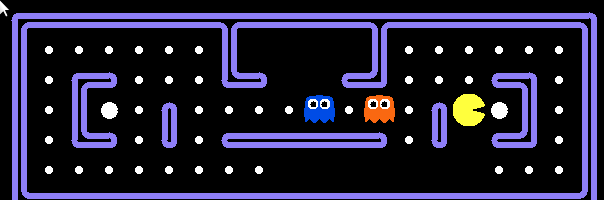}
    \vspace{-0.2cm}
    \caption{Small PAC-Man environment}
    \label{fig:small_pacman_environment}
    \end{figure}
% We illustrate concepts through PAC-Man throughout this paper.
Figure~\ref{fig:small_pacman_environment} shows a small PAC-Man environment from~\cite{berkeley_pacman}, which serves as a running example. 
The agent can move in the four cardinal directions or do nothing at every time step. Its goal is to eat all pellets (small dots). Colliding with ghosts terminates an episode unless they are vincible, which happens for a fixed amount of time after the agent eats a capsule (large dot). The observable states include information on the location of walls, the distance to ghosts, capsules, and pellets, the direction towards these objects, and other data. For efficient learning, we use a one-hot encoding for categorical features, e.g., features $9$ to $12$ describe the direction toward the closest pellet, where $f_9^s = 1$ means that the closest pellet is north.   

The agent receives a reward of $10$ for eating a pellet, $200$ for eating a ghost (collision when vincible), $500$ for completing the level, $-500$ for colliding with an invincible ghost, and $-1$ every time step to encourage fast completion.

\section{Step 1: Rule Mining}
\label{sec:mine_rules}
To create training data for rule mining, we execute the agent in its environment for $n_{rule}$ episodes to sample traces, i.e., state-action-reward sequences. % from the induced Markov chain $\mathcal{M}^\pi$. 
From these traces, we collect all observed state-action pairs $(s,a) \in S \times A$ in a multiset $E$ to which we refer as experiences. 
%Let $E$ be the multiset containing all such experiences. 
We treat rule mining as a classification problem from states to actions based on training data $E$, where we aim to learn rules that (1) represent what is important to the agent's decisions and that (2) are accurate and have high coverage. In the following, we detail the form of rules we consider and how we learn them.

\noindent
\textbf{Rules.}
A rule $\rlrule$ is of the form:
\begin{equation}
   \rlrule = \otimes action(a) \leftarrow \bigwedge_{i \in I} f_i = v_i, \text{ where}
\end{equation}
 $\otimes \in \{+,-\}$ is the rule polarity, and $I \subseteq [1..n]$ is a set of (feature) indices. We refer to the conjunction of conditions as the rule body, denoted $body(\rlrule)$, and to the action consequent as the rule head. 
Given a state $s$ which satisfies $\bigwedge_{i \in I} v_i = f_i^s$, we say that $\rlrule$ triggers in $s$, denoted $s \models \rlrule$. If $\otimes = -$, $\rlrule$ is a negative rule, denoting action $a$ is avoided if $s \models \rlrule$. A positive rule with $\otimes = +$ denotes that $a$ should be taken. 

\noindent
\textbf{Feature Importance.}
To determine which features are important to the policy's decision, we select  $n_{feat}$ experiences from $E$ and apply \textsc{Lime} to generate explanations for the decisions of $\pi$. The explanations quantify the importance of a feature $f_i$ to perform or not perform an action $a$. For each $f_i$, we compute the sum of importance values from the $n_{feat}$ explanations, which we denote by $imp(\otimes, a, f_i)$, representing the importance of feature $f_i$ to take or avoid $a$.

\noindent
\textbf{Mining Rules.}
We apply the rule learning algorithm \mbox{\textsc{Ripper}}~\cite{DBLP:conf/icml/Cohen95}
for every combination of action and rule polarity individually. This approach enables more effective mining of negative rules compared to posing rule mining as a multi-class classification problem for all actions. Hence, we select two datasets from $E$ for every action-polarity combination, which rules shall distinguish: the inclusion dataset $inc(a,\otimes)$ containing positive examples and the exclusion dataset $exc(a,\otimes)$. To account for feature importance, we adapt \textsc{Ripper}'s rule growing. Instead of solely targeting coverage by optimizing FOIL's information gain during rule growing, we maximize the product of FOIL's information gain and feature importance $imp(\otimes, a, f_i)$.

The inclusion dataset $inc(a,+)$ for mining positive rules contains all $(s,a)$ from $E$, i.e., experience matching action $a$, and the exclusion dataset contains all $(s,a')$ from $E$ s.t. $a' \neq a$. To mine negative rules, we add all $(s,a')$ where $a' = \argmin_a Q(s,a)$ in $inc(a,+)$ and we add all $(s,a)$ to $exc(a,-)$.
% for positive rules are given by $inc(a,+) = \{ s \mid (s,a) \in E\}$ and $exc(a,+) \{s  \mid (s,a') \in E \land a' \neq a\}$ 
% and for negative rules, they are $inc(a,-) = \{s \mid (s,a') \in E \land a' = \argmin_a Q(s,a) \}$ and $exc(a,-) = \{ s \mid (s,a) \in E\}$. After determining the sets, we apply \textsc{Ripper} to distinguish the inclusion and exclusion sets.
Following the rule mining through \textsc{Ripper}, we filter rules by imposing bounds on minimal accuracy and coverage.
We evaluate the accuracy of a rule by checking if it agrees with the experiences from a validation set. Coverage refers to the ratio of times that a rule triggers, which is a common optimization criterion of rule learning algorithms.

\noindent
\textbf{Running Example.}
In the PAC-Man environment, we learn rules, such as $\rlrule_1 = -action(0)\leftarrow f_{62}=0$ and $\rlrule_2 = +action(0)\leftarrow f_9=1\land f_{50} =1 \land f_{62}=1 \land f_{65}=1$. Both predicate on action $0$, i.e., going north. Rule $\rlrule_1$ says that PAC-Man avoids trying to go north if there is a wall, as $f_{62}=0$ means that going north is not possible. The second rule $\rlrule_2$ says that the agent learned to go north if there is a pellet in the north direction ($f_9=1$), Ghost 2 is west ($f_{50} =1$), and there is no wall to the north ($f_{62} =1$) and west ($f_{65}=1)$.

\section{Step 2: Rule Generalization}
The rules provide insights into decisions that the policy learned to be beneficial. The intuition behind rule generalization is that often there are symmetries in the environment and the agent may have learned how to act in one situation, but not in corresponding symmetrical situations. More generally, situations are often related through relations that can be expressed symbolically. Borrowing the concept from software testing, we refer to these symbolic relations as metamorphic relations (MRs). We define MRs for RL based on feature relations that relate actions and individual features of states.
% A feature relation $\fr$ is a tuple $\langle a,a', f_i=x,f_j=y\rangle$ describing a relation between state-action pairs. Two state-action pairs $(s_1,a_1)$ and $(s_2,a_2)$ are feature-related by $\fr$ denoted $(s_1,a_1) \fr (s_2,a_2)$ iff
% $a_1 = a \land a_2 = a' \land f_i^{s_1} = x \land f_j^{s_2} = y$. A metamorphic relation $\mr$ for RL is a set of feature relations defined on the same actions $a$ and $a'$. 
% Given  $sa_1 = (s_1,a_1)$ and $sa_2 = (s_2,a_2)$, we define $sa_1 \mr sa_2$ iff $sa_1 \fr sa_2$ for all $\fr \in \mr$.\todo{revise, for all makes sense for rules but not for states}
\begin{definition}
A feature relation $\fr$ is a tuple $\langle a,a', i,j,\mathcal{R}\rangle$ with $\mathcal{R}\subseteq \mathbb{R} \times \mathbb{R}$ describing a relation between features $i$ and $j$ in state-action pairs. Two state-action pairs $(s_1,a_1)$ and $(s_2,a_2)$ are feature-related by $\fr$ denoted $(s_1,a_1) \fr (s_2,a_2)$ iff
$a_1 = a \land a_2 = a' \land (f_i^{s_1},f_j^{s_2}) \in \mathcal{R}$. A metamorphic relation $\mr$ for RL is a set of feature relations defined on the same actions $a$ and $a'$. 
Given  $sa_1 = (s_1,a_1)$ and $sa_2 = (s_2,a_2)$, we define $sa_1 \mr sa_2$ iff $sa_1 \fr sa_2$ for all $\fr \in \mr$, i.e., an MR relates multiple pairs of features of state-action pairs.
\end{definition}

\begin{figure}[t]
    \centering
    \begin{tikzpicture}
    
    \node(east_wall){\includegraphics[width=0.12\linewidth]{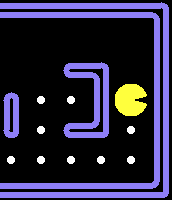}};
    \node[below right=0cm and 0.4cm of east_wall.east,single arrow, draw, fill=black!20, 
      minimum width = 10pt, single arrow head extend=3pt, rotate = 90,
      minimum height=7mm](go_north){};
      \node[below = 0.3cm of go_north]{North};

    \node[right = 1.5 cm of east_wall,draw,fill=blue!20](fr){$\fr$};
    
    \node(north_wall)[right= 2.5 cm of east_wall]{\includegraphics[width=0.12\linewidth]
    {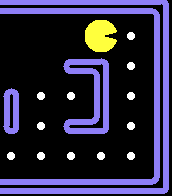}};

    \node[circle,draw, fill=white, inner sep=1pt, font=\scriptsize]at([shift=({0.3cm,-0.3cm})]north_wall.north west){$s^2$};
    \node[circle,draw, fill=white, inner sep=1pt, font=\scriptsize]at([shift=({0.3cm,-0.3cm})]east_wall.north west){$s^1$};
    
    \node[right=0.9cm of north_wall.east,single arrow, draw, fill=black!20, 
      minimum width = 10pt, single arrow head extend=3pt, rotate = 180,
      minimum height=7mm](go_east){};
      \node[below = 0.3cm of go_east]{East};

    \node[font=\Huge, left = 0cm of east_wall]{$($};
    \node[font=\Huge, left = 0cm of north_wall]{$($};
    \node[font=\Huge, right = 1cm of east_wall]{$)$};
    \node[font=\Huge, right = 1cm of north_wall]{$)$};
    \node[font=\Huge, below right = -0.9cm and -0.2cm of east_wall]{$,$};
    \node[font=\Huge, below right =-0.9cm and -0.2cm of north_wall]{$,$};
    \end{tikzpicture}
    \vspace{-0.3cm}
    \caption{State-action pairs related by $\fr = \langle 0,2, 62,64, = \rangle$.}
    \label{fig:feature_rel}
\end{figure}
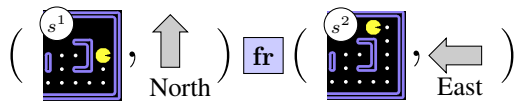
\noindent
\textbf{Running Example.}
Figure~\ref{fig:feature_rel} shows two PAC-Man state-action pairs related by $\langle 0,2, 62,64, = \rangle$, where $=$ simply denotes the equality relation. On the left, PAC-Man goes north action with action $0$ and there is no wall to the north, represented by $f_{62}^\mathit{s^1}=1$. On the right, PAC-Man goes east with action $2$ and there is no wall to the east $f_{64}^\mathit{s^2}=1$, i.e., we have $f_{62}^\mathit{s^1} = f_{64}^\mathit{s^2}$ satisfying the relation from $\fr$.

We define MRs based on feature relations to enable  convenient, compositional definitions and
% Feature relations further enable a simpler way to account for vision-based input
extend them to rules as follows. 
Let $\fr=\langle a,a', j,k,\mathcal{R}\rangle$, $\rlrule_1 = \otimes_1 action(a_1) \leftarrow \bigwedge_{i \in I_1} f_i = {v_{i1}}$, and $\rlrule_2 = \otimes_2 action(a_2) \leftarrow \bigwedge_{i \in I_2} f_i = {v_{i2}}$: 
%$\rlrule_1$ and $\rlrule_2$ are feature related denoted 
$\rlrule_1 \fr \rlrule_2$ iff $\otimes_1 = \otimes_2$, $a_1 = a$, $a_2 = a'$, and either $f_j=x \in body(\rlrule_1)$ and $f_k = y \in body(\rlrule_2)$ such that $(x,y) \in \mathcal{R}$  or $j \notin I_1$ and $k \notin I_2$. Rule bodies must either include facts satisfying the relation $\mathcal{R}$, or they should not predicate on the respective features. 
Analogously to states, we extend MRs to rules denoted by $\rlrule_1 \mr \rlrule_2$.
Through sets of MRs $M$, we automatically generalize each positive and negative rule $\rlrule_j$ to a set of positive or negative rules given by the fixed point of $M_{gen}(\rlrule_j) = \{\rlrule_j\} \cup \bigcup\{M_{gen}(\rlrule) \mid \rlrule_j \mr \rlrule, \mr \in M\}$. 

\noindent
\textbf{Running Example.}
We illustrate rule generalization with rule $\rlrule_1 = -action(0)\leftarrow f_{62}=0$ and feature relation $\fr = \langle 0,2, 62,64, = \rangle$ from above.  % is relevant for $\rlrule_1$. 
An MR $\mr = \{\fr\}$ including only $\fr$ describes a 90-degree clockwise rotation of the action and the only relevant feature from north to east. Action $0$ and $2$ describe going north and east, respectively. Features $62$ and $64$ indicate the absence of a wall to the north and east, respectively, e.g. $f_{64}=0$ means that there is a wall to the east. Hence, if $\rlrule_2 = -action(2)\leftarrow f_{64}=0$ then $\rlrule_1 \mr \rlrule_2$, because $\rlrule_1 \fr \rlrule_2$, which holds because $polarity(\rlrule_1) = polarity(\rlrule_2)$, $head(\rlrule_1) = 0$, $head(\rlrule_2) = 2$, and $f_{62}=0 \in body(\rlrule_1)$,
$f_{64}=0 \in body(\rlrule_2)$ with $0=0$. The rule $\rlrule_2$ can be automatically generated and states to not go east if there is a wall. Enforcing it may improve performance if the agent has not learned to generalize from $\rlrule_1$ to $\rlrule_2$, because trying to go into a wall results in a reward of $-1$ as the agent stays in its location in such a case.

\section{Step 3: Guiding Agents with Rules}
To evaluate $\pi$, we propose to first execute $\pi$ without rule guidance and then guide $\pi$ with each rule set $M_{gen}(\rlrule_j)$ separately for a set $M$ of MRs. By comparing the gained cumulative reward, we 
determine which generalized rules improve performance and thus reveal weaknesses of $\pi$. 

\begin{algorithm}[t]
\caption{Policy evaluation guided by Rules}
\label{alg:eval}
\footnotesize
\begin{algorithmic}[1]
\REQUIRE Q-function $Q$, set of gen. rules $G$, \# eval. episodes $n$
\ENSURE Average Cumulative Reward
\STATE $Rews \gets \langle \rangle$
\FOR{$i \gets 1 \textbf{ to }n$}
\STATE $s \gets \textsc{Reset()}$, $rew \gets 0$
\WHILE{$s$ not terminal}
\STATE $G_t \gets \{(action(r),polarity(r)) \mid r\in G,s \models r\}$
\label{algline:triggers}
\IF{$|\{(a,+) \in G_t\}| = 1$} \label{algline:enforce_if}
\STATE $act \gets a$
\ELSE %\COMMENT $\{(a,+) \in G_t\}| \neq 1$
\STATE $q \gets Q(s, \cdot)$
\STATE $q(s,a) \gets -\infty \text{ for } (a,-) \in G_t$\label{algline:neg_start}
\STATE $act \gets \argmax_a q(s,a)$
\ENDIF
\STATE $s,r \gets \textsc{Step}(s,act)$, $rew \gets rew + r$
\ENDWHILE
\STATE $\textsc{Append}(Rews,rew)$
\ENDFOR
\RETURN $mean(Rews)$, $stderr(Rews)$
\end{algorithmic}
\end{algorithm}

Algorithm~\ref{alg:eval} monitors the execution of $\pi$ for $n$ episodes 
and enforces rules if they trigger. It assumes that the environment provides $\textsc{Reset}$ and $\textsc{Step}$ operations, which are commonly part of the interface to RL environments, like the Gymnasium API~\cite{towers2024gymnasiumstandardinterfacereinforcement}. After resetting the environment to start an episode, in every step, Line~\ref{algline:triggers} checks which rules trigger. If a single positive rule triggers (Line~\ref{algline:enforce_if}), we enforce it. Otherwise -- there is no positive rule or a conflict between positive rules -- we disable all actions of negative rules in Line~\ref{algline:neg_start} by setting their Q-values to negative infinity. After that, we choose the action with the highest Q-value or a random action if all actions have been disabled.  Furthermore, we could add additional randomness to the action choices. 
Finally, Algorithm~\ref{alg:eval} returns the average cumulative reward and the corresponding standard error. 
While generalization does not change rule polarity, Algorithm~\ref{alg:eval} supports combining rules generalized from rules with different polarity.

\section{Experiments}

This section presents experiments on the application of learned and generalized rules. First, we show how generalized rules help detect weaknesses in RL policies through rule-guided execution and how rules can explain the identified weaknesses. After that, we demonstrate the RL policy improvement through rule-guided execution. The appendix includes an example on the application of rules to pinpoint and explain the reason for observed policy behavior.

\noindent
\textbf{Setup and Environment.}
All experiments are based on RL policies trained in six PAC-Man levels~\cite{berkeley_pacman} and five \texttt{highway-env}~\cite{highway_env} environments using stable-baselines3~\cite{sb3}. The PAC-Man levels differ in size (small, medium, and original) and the presence of capsules. 
% The states observed by the RL agent are  in the small and medium environments and $117$-dimensional vectors in the original-sized environments. 
We trained DQN~\cite{DBLP:journals/nature/MnihKSRVBGRFOPB15} policies for $2.5\cdot 10^6$ steps in the small and medium PAC-Man environments with $69$-dimensional states and for $5 \cdot 10^6$ steps in the original-sized environments with $117$-dimensional states. In  \texttt{highway-env}, we trained DQN policies for $5 \cdot 10^5$ steps to navigate in driving scenarios, like merging onto a highway. We configured \texttt{highway-env} to create observations relative to the ego vehicle, comprising $7$ properties of the ten closest vehicles, like their relative positions. Every training run is repeated five times and the resulting policies provide the basis for the experiments below. Code and data from the experiments are available at \url{https://doi.org/10.6084/m9.figshare.28569017}.

We mine rules for all environments using the same setup, except for the discretization of states. In the PAC-Man environments, we use the \texttt{decile} discretization provided by \textsc{Lime}~\cite{python_lime}, which discretizes each numerical feature into intervals corresponding to deciles calculated from the data. Categorical features, like cardinal directions to the closest food, are left unchanged. 
The \texttt{highway-env} environments do not include categorical features and there we normalize and discretize each feature into ten intervals.  
For both types of environments, we sample $n_{rule} = 600$ episodes to generate experiences for rule mining and impose a minimal accuracy of $0.9$ and a minimal coverage of $0.01$ on rules, discarding all other rules.

For the remainder of this section, let $\pi$ denote a trained policy, $R$ be a set of rules learned from $\pi$, and $M$ be a set of MRs. The MRs for PAC-Man encode 90-degree clockwise turns from every cardinal direction, i.e., from every rule, we generate four rules, including the original rule. The \texttt{highway-env} MRs generalize rules to other vehicles occurring in observations, i.e., from a rule with conditions on the first vehicle, we generate rules conditioned on the other vehicles. Furthermore, the MRs encode symmetry constraints, e.g. generalizing rules for changing to the left lane to rules for changing to the right lane and vice versa.

\subsection{Identifying Weaknesses}
With the first set of experiments, we approach the research question RQ1: \emph{Does rule-guided execution effectively reveal weaknesses in a trained policy?} We consider policy weaknesses as decisions where enforcing a rule improves performance, i.e., rule-guided execution detects decisions that could be improved. As a benchmark for performance, we compare against the average cumulative reward $cr_\pi$ gained by $\pi$ without rule guidance.
To identify weaknesses in $\pi$, we propose to generalize each rule $r \in R$ to $R_g = M_{gen}(r)$ individually and guide the execution of $\pi$ with $R_g$ using Algorithm~\ref{alg:eval}. We perform $n = 100$ evaluation episodes in \texttt{highway-env} and $n=250$ evaluation episodes in the PAC-Man environments. $R_g$ reveals a weakness if the cumulative rewards from rule-guided execution are significantly larger than those of unguided execution, which we determine using a Welch test~\cite{welch} and a p-value threshold of $0.05$. 

\noindent
\textbf{Baselines.}
We are the first to propose MT for RL, therefore we compare MT to two random baselines. The first baseline, \emph{random testing (RT)}, blocks or enforces randomly chosen actions during policy execution in $k$ percent of the states. 
% where we randomly choose $k$ sets of actions to be blocked or actions to be enforced randomly. 
We randomly assign each of the $k$ policy changes to approximately one percent of the state space, such that if we visit a state twice the same change happens. In the experiments, we performed $100$ RT evaluations where we only blocked actions and $100$ evaluations where only enforced actions. We set $k=3$ since we found that low values are more effective at detecting weaknesses. Each evaluation compares against the average cumulative reward $cr_\pi$ from unguided execution. 

Additionally, we use another random baseline, which we call \emph{random rules} (RR). For RR, we randomly generate rule sets instead of generating them via MRs and execute Algorithm~\ref{alg:eval}, i.e., RR serves as an \textbf{ablation study} to study the impact of MRs. The random generation uses information extracted from mined rules by creating rules with the same distribution of rule lengths and rule polarities, and by using the feature values found in the mined rules. Additionally, rule sets created for RR are of similar size as the rule sets created using MRs. Hence, the random rules benefit from information extracted from policies, but not from domain knowledge. 

% If the resulting mean cumulative reward $cr_{R_g}$ is larger than $cr_\pi$ from non-guided execution of $\pi$, we assume to have identified a weakness. It shows that $\pi$ has not learned an advantageous decision included in $R_g$.
% We further distinguish \emph{uncertain} and \emph{certain}  weaknesses. 
% Let $r$ and $r_g$ be the average cumulative reward from non-guided and guided execution with a rule set $R_g$ via Algorithm~\ref{alg:eval} and let $err$ and $err_g$ be the corresponding standard error. 
% We say that $R_g$ reveals an uncertain weakness if $cr_g > cr_\pi$ and that it reveals a certain weakness if $cr_g - err_g > cr_\pi + err$, where $err_g$ and $err$ standard errors returned by Algorithm~\ref{alg:eval}.  

\begin{table}[t]
\centering
\footnotesize
    
         \caption{Detected weaknesses and number of evaluations in six PAC-Man levels.}
         
              \vspace{-0.3cm}
       \begin{tabular}{l|c|c|c|c}
         \multirow{2}{*}{Experiment} &  \multicolumn{3}{|c|}{Detected Weaknesses} &  \ \multirow{2}{*}{\# Evaluations} \\ \cline{2-4}
          & RT & RR & MT (Ours) &  \\ \hline 
           small & $0.00$ & $0.01$ & $\mathbf{0.13}$ & $105.4$ \\ \hline 
 small-nc & $0.01$ & $0.03$ & $\mathbf{0.11}$ & $85.0$ \\ \hline 
 medium & $0.01$ & $0.04$ & $\mathbf{0.08}$ & $117.6$ \\ \hline 
 medium-nc & $0.01$ & $0.00$ & $\mathbf{0.09}$ & $109.6$ \\ \hline 
 original & $0.00$ & $0.00$ & $\mathbf{0.16}$ & $121.6$ \\ \hline 
 original-nc & $0.02$ & $0.03$ & $\mathbf{0.14}$ & $108.8$ \\  

         \end{tabular}

    \label{tab:weaknesses-pac}
\end{table}

\begin{table}[t]
\footnotesize
\centering
 
              \caption{Detected weaknesses and number of evaluations in five \texttt{highway-env} environments.}
              
              \vspace{-0.3cm}
         \begin{tabular}{l|c|c|c|c}
         \multirow{2}{*}{Experiment} &  \multicolumn{3}{|c|}{Detected Weaknesses} &  \ \multirow{2}{*}{\# Evaluations} \\ \cline{2-4}
          & RT & RR & MT (Ours) &  \\ \hline 
           highway & $0.02$ & $\mathbf{0.14}$ & $0.05$ & $90.8$ \\ \hline 
 highway-fast & $0.07$ & $\mathbf{0.20}$ & $0.12$ & $119.6$ \\ \hline 
 merge & $0.09$ & $\mathbf{0.24}$ & $0.16$ & $109.6$ \\ \hline 
 intersection & $0.06$ & $0.00$ & $\mathbf{0.12}$ & $20.8$ \\ \hline 
 roundabout & $0.01$ & $0.06$ & $\mathbf{0.11}$ & $33.6$ \\  

         \end{tabular}
         
    \label{tab:weaknesses-hw}
\end{table}

\noindent
\textbf{Results.}
Tables~\ref{tab:weaknesses-pac} and ~\ref{tab:weaknesses-hw} show evaluation results for the five base policies trained in each environment. For each RT, RR, and rule-guided execution supported by MRs, denoted as metamorphic testing (MT), the tables show the mean ratio of evaluations that detected weaknesses from five repetitions. The tables additionally show the number of evaluations for RR and MT (for RT they are fixed to two times $100$), i.e., the number of generalized rule sets $R_g$. We can see that on average $5$ to $24$ percent of the MT evaluations reveal weaknesses. This is consistently higher than RT, especially in PAC-Man environments, where RT rarely reveals weaknesses. In contrast, RR reveals more weaknesses than MT in three cases where it substantially lees weaknesses in the other eight environments. 
% In only three environments (highway, highway-fast, and intersection), at least one evaluation does not reveal a weakness, as the corresponding minimum ratio equals zero. 
We can answer RQ1 positively, as rule-guided execution consistently revealed weaknesses in the examined policies, both through RR and MT. We can further deduce that domain knowledge helps since MT performed better overall. In the cases, where RR performed better than MT, either our MRs may not optimally represent relations that hold in the environment or the policy under test generalizes well, but includes other issues that RR detects. MT further improves upon RR as it facilitates explanations, as the rules within a rule set are related. If a rule set $R_g$ reveals weakness, we know that the cause is linked to the original rule from which $R_g$ was generated and to the MRs used for generation. Beyond that, the good performance of RR suggests that search-based approaches, which are popular in RL testing~\cite{DBLP:journals/tse/ZolfagharianABBS23,DBLP:conf/ijcai/TapplerCAK22}, might be viable for rule generation. 

% In each of the six environments, we identified at least one set of generalized $R_g$ that improves, thus we can answer the above research question positively.

\paragraph{Explaining Weaknesses.}
We illustrate explaining policy weaknesses with two cases from the PAC-Man levels \emph{small} and \emph{original} with capsules. In both cases, the generalization of the following simple rules increases performance:
\begin{align*}
-action(0) &\leftarrow f_9=0 \land f_{62}=0 \tag{small} \\
-action(0) &\leftarrow f_9=0 \land f_{110}=0 \tag{original}
\end{align*}
The feature indices differ among the environments, but both rules can be interpreted as "\emph{don't go north if there is no food and there is a wall to the north}". Generalizations to other directions are safe to apply, yet the policies have not picked up on them, thus enforcing them improves performance. However, we found that the policies have learned stronger versions of these rules, e.g., in the environment \emph{original}, we learned related rules with additional constraints: $-action(1) \leftarrow f_2=0\land f_9=0\land f_{10}=0\land f_{11}=1\land f_{88}=0\land f_{111}=0$ and $-action(3) \leftarrow f_5=0\land f_{10}=1\land f_{12}=0\land f_{113}=0$. Hence, rule-guided execution points to situations where the policy learned suboptimal decisions. 

\begin{algorithm}[t]
\footnotesize
\begin{algorithmic}[1]
\STATE $cr_{max} \gets \text{mean cumulative reward of } \pi$
 \STATE $RT \gets \emptyset$ 
 \FOR{$\rlrule \in R$}
\IF{Alg. \ref{alg:eval} with $G \gets M_{gen}(\rlrule) = R_g$ reveals weakness}
\STATE $RT' \gets  RT \cup R_g$
\STATE $cr,stderr \gets \text{Algorithm~\ref{alg:eval} with } G \gets RT'$
\IF{$cr > cr_{max}$}
\STATE $RT\gets RT'$, $cr_{max} \gets cr$
\ENDIF
\ENDIF
 \ENDFOR
\end{algorithmic}
\caption{Greedy rule selection for policy improvement.}
\label{alg:greedy}
\end{algorithm}
\subsection{Policy Improvements}
The final set of experiments focuses on policy improvement without retraining through rule-guided execution, where we tackle Research Question RQ2: \emph{Can guidance with compositions of rule sets improve RL policies?} For this purpose, we assume to have already evaluated $\pi$ with all generated sets of rules $R_g$ individually. To improve $\pi$, we propose to compose rule sets (RS) that reveal weaknesses without overconstraining the policy. Below, we present experiments, where we first greedily selected rule compositions $RT$ through Algorithm~\ref{alg:greedy}.

\begin{table}[t]
\footnotesize
 %                 \begin{tabular}{l|l|l|l|l}
 %                 Experiment &  Base & Rule-Guided (Ours) & Extended Training & \# Rule Sets  \\ \hline
 %                   small & $1023.59 \pm 43.68$ & $1268.91 \pm 20.53$ & $1018.41 \pm 14.02$ & $4.60$ \\ \hline 
 % small(NC) & $450.61 \pm 36.52$ & $635.71 \pm 16.76$ & $447.93 \pm 11.64$ & $4.40$ \\ \hline 
 % medium & $1339.58 \pm 48.97$ & $1601.12 \pm 21.96$ & $1371.39 \pm 14.94$ & $4.40$ \\ \hline 
 % medium(NC) & $840.90 \pm 38.60$ & $1050.23 \pm 16.95$ & $789.45 \pm 12.41$ & $3.80$ \\ \hline 
 % original & $1590.55 \pm 51.48$ & $2013.56 \pm 28.84$ & $1552.42 \pm 16.27$ & $4.20$ \\ \hline 
 % original(NC) & $784.75 \pm 48.31$ & $1194.81 \pm 28.11$ & $796.44 \pm 15.02$ & $3.40$ \\  

 %                 \end{tabular}
              \caption{Average cumulative reward and standard error for non-guided and rule-guided execution of PAC-Man.}
              \vspace{-0.3cm}
                         \begin{tabular}{l|c|c|c|c}
                 Experiment &  Base & \makecell{Rule-Guided \\ (Ours)} & Ext. Training & RS  \\ \hline
                   small & \makecell{$1023.6$ \\ $\pm 43.7$} & \makecell{ $\mathbf{1268.9}$ \\ $\mathbf{\pm 20.5}$} & \makecell{ $1018.4$ \\ $\pm 14.0$ } & $4.6$ \\ \hline 
 small-nc & \makecell{$450.6$ \\ $\pm 36.5$} & \makecell{ $\mathbf{635.7}$ \\ $\mathbf{\pm 16.8}$} & \makecell{ $447.9$ \\ $\pm 11.6$ } & $4.4$ \\ \hline 
 medium & \makecell{$1339.6$ \\ $\pm 49.0$} & \makecell{ $\mathbf{1601.1}$ \\ $\mathbf{\pm 22.0}$} & \makecell{ $1371.4$ \\ $\pm 14.9$ } & $4.4$ \\ \hline 
 medium-nc & \makecell{$840.9$ \\ $\pm 38.6$} & \makecell{ $\mathbf{1050.2}$ \\ $\mathbf{\pm 16.9}$} & \makecell{ $789.4$ \\ $\pm 12.4$ } & $3.8$ \\ \hline 
 original & \makecell{$1590.5$ \\ $\pm 51.5$} & \makecell{ $\mathbf{2013.6}$ \\ $\mathbf{\pm 28.8}$} & \makecell{ $1552.4$ \\ $\pm 16.3$ } & $4.2$ \\ \hline 
 original-nc & \makecell{$784.8$ \\ $\pm 48.3$} & \makecell{ $\mathbf{1194.8}$ \\ $\mathbf{\pm 28.1}$} & \makecell{ $796.4$ \\ $\pm 15.0$ } & $3.4$ \\  

                 \end{tabular}
    \label{tab:improvements-pac}
\end{table}

\begin{table}[t]
\footnotesize
              \caption{Average cumulative reward and standard error for non-guided and rule-guided execution of policies in \texttt{highway-env}.}
              \vspace{-0.3cm}

                 \begin{tabular}{l|c|c|c|c}
                 Experiment &  Base & \makecell{Rule-Guided \\ (Ours)} & Ext. Training & RS  \\ \hline
                   highway & \makecell{$41.7$ \\ $\pm 3.3$} & \makecell{ $\mathbf{110.4}$ \\ $\mathbf{\pm 9.7}$} & \makecell{ $68.9$ \\ $\pm 5.1$ } & $5.4$ \\ \hline 
 highway-fast & \makecell{$48.1$ \\ $\pm 4.5$} & \makecell{ $\mathbf{193.8}$ \\ $\mathbf{\pm 14.1}$} & \makecell{ $74.3$ \\ $\pm 5.2$ } & $3.6$ \\ \hline 
 merge & \makecell{$23.8$ \\ $\pm 0.3$} & \makecell{ $\mathbf{28.7}$ \\ $\mathbf{\pm 0.1}$} & \makecell{ $25.6$ \\ $\pm 0.3$ } & $2.8$ \\ \hline 
 intersection & \makecell{$11.5$ \\ $\pm 0.7$} & \makecell{ $\mathbf{12.7}$ \\ $\mathbf{\pm 0.5}$} & \makecell{ $11.7$ \\ $\pm 0.5$ } & $2.0$ \\ \hline 
 roundabout & \makecell{$361.0$ \\ $\pm 45.4$} & \makecell{ $\mathbf{777.2}$ \\ $\mathbf{\pm 21.4}$} & \makecell{ $499.0$ \\ $\pm 32.9$ } & $4.6$ \\  

                 \end{tabular}
    \label{tab:improvements-hw}
\end{table}

Table~\ref{tab:improvements-pac} and Table~\ref{tab:improvements-hw} show the cumulative reward of non-rule-guided execution and execution guided by $RT$. We performed the non-rule-guided execution with the base policies from which we mined rules and with policies trained twice as long, e.g., policies for \texttt{highway-env} were trained for $10^6$ steps. 
% For this purpose, we trained each base policy another time for as long as it has been trained originally, i.e., instead of training for $5\cdot 10^6$ steps in the original PAC-Man environment, we trained for a combined $1 \cdot 10^7$ steps. 
The tables again report averages from five repetitions, particularly the mean cumulative reward from running Algorithm~\ref{alg:eval} and the standard error of the estimate of the mean cumulative reward. The tables also show how many rule sets have been composed to create $RT$.
We can see that rule-guided execution always improved the base reward substantially, except in the \emph{intersection} environment, where the improvement is only marginal on average. In other cases, we see larger improvements, e.g., $274$ percent in the \emph{highway-fast} environment. We can further see that extended training did not improve the reward in many environments (PAC-Man) and only slightly in others (\texttt{highway-env}). In all cases, rule-guided execution yields larger improvements, thus it provides value that cannot be gained from more training. Moreover, the greedy selection of rules always composed multiple rule sets to create $RT$ to improve the policy under consideration. Hence, we can answer RQ2 positively.

\section{Conclusion}
We propose \algname{}, a rule-based framework for evaluating, explaining, and improving RL policies. After learning rules from the most important policy decisions, we use high-level domain knowledge to generalize learned rules to other related situations. For this purpose, we leverage MRs~\cite{DBLP:journals/csur/ChenKLPTTZ18} to express relations like symmetries that hold in the considered environment. By executing policies guided by generalized rules, we identify weaknesses, where a policy learned to behave adequately in a particular situation, but not in related situations. That is, we find cases where RL policies fail to generalize. In these cases, the generalized rules and applied MRs explain the found weaknesses.
\algname{} provides a basis for policy improvement, by enforcing sets of generalized rules, which were found to improve performance. In experiments with deep RL policies trained in six PAC-Man environments~\cite{berkeley_pacman} and in five \texttt{highway-env}~\cite{highway_env} environments, \algname{} revealed weaknesses in all policies, which are explained by the learned and generalized rules. Rule-guided execution improved the average cumulative reward by up to $273 \%$.

\algname{} enables the integration of domain knowledge into RL policies and our evaluation approach is the first metamorphic testing (MT) approach for RL. In contrast to existing work on RL testing~\cite{DBLP:conf/ijcai/TapplerCAK22,DBLP:journals/tse/ZolfagharianABBS23,DBLP:conf/kbse/LiWZCCZXMZ23,DBLP:journals/tosem/BiagiolaT24}, which brings the agent into challenging environment states, we take an agent-centric view, evaluating changes in the agent's decisions. By generalizing decisions learned from a policy, we ensure that generalized decisions are learnable.

In future work, we will investigate how to integrate other types of domain knowledge, for example, knowledge about temporal dependencies between actions via restraining bolts~\cite{DBLP:conf/aaai/GiacomoIFP20}. Since our work is complementary to existing RL testing approaches, we will study how to combine MT with existing work that focuses on the environment, like search-based testing~\cite{DBLP:conf/ijcai/TapplerCAK22,DBLP:journals/tse/ZolfagharianABBS23,DBLP:journals/tosem/BiagiolaT24}. Finally, we will work on a more seamless integration of generalized rules into RL policies through rule-guided training.

\appendix
% Authors can include in the main body of their paper, or on the reference pages, an ethics statement that addresses both ethical issues regarding the research being reported and the broader ethical impact of the work. Note that such an ethics statement is not required, but we recommend that papers working with sensitive data or on sensitive tasks include such discussion. The IJCAI review form will include a section asking reviewers and ACs to flag any serious ethical concerns.

% \section*{Ethical Statement}

% There are no ethical issues.

\section*{Acknowledgments}
% The work by M. Tappler, I. D. Lopez-Miguel, and E. Bartocci has been funded by the Vienna Science and Technology Fund (WWTF), projects [10.47379/ICT22-023] and [10.47379/ICT19018]. The work by S. Tschiatschek has been funded by the Vienna Science and Technology Fund (WWTF) [10.47379/ICT20058].
This work has been funded by the Vienna Science and Technology Fund (WWTF) [10.47379/ICT22-023], [10.47379/ICT19018], [10.47379/ICT20058]. 

% TODO

\bibliographystyle{named}
%\newpage
\bibliography{ijcai24}

\begin{thebibliography}{}

\bibitem[\protect\citeauthoryear{Alshiekh \bgroup \em et al.\egroup }{2018}]{DBLP:conf/aaai/AlshiekhBEKNT18}
Mohammed Alshiekh, Roderick Bloem, R{\"{u}}diger Ehlers, Bettina K{\"{o}}nighofer, Scott Niekum, and Ufuk Topcu.
\newblock Safe reinforcement learning via shielding.
\newblock In {\em AAAI}, pages 2669--2678. {AAAI} Press, 2018.

\bibitem[\protect\citeauthoryear{Apt{\'{e}} and Weiss}{1997}]{DBLP:journals/fgcs/ApteW97}
Chidanand Apt{\'{e}} and Sholom~M. Weiss.
\newblock Data mining with decision trees and decision rules.
\newblock {\em Future Gener. Comput. Syst.}, 13(2-3):197--210, 1997.

\bibitem[\protect\citeauthoryear{Bastani \bgroup \em et al.\egroup }{2018}]{DBLP:conf/nips/BastaniPS18}
Osbert Bastani, Yewen Pu, and Armando Solar{-}Lezama.
\newblock Verifiable reinforcement learning via policy extraction.
\newblock In {\em NeurIPS}, pages 2499--2509, 2018.

\bibitem[\protect\citeauthoryear{Bewley and Lawry}{2021}]{DBLP:conf/aaai/BewleyL21}
Tom Bewley and Jonathan Lawry.
\newblock Tripletree: {A} versatile interpretable representation of black box agents and their environments.
\newblock In {\em AAAI}, pages 11415--11422. {AAAI} Press, 2021.

\bibitem[\protect\citeauthoryear{Biagiola and Tonella}{2024}]{DBLP:journals/tosem/BiagiolaT24}
Matteo Biagiola and Paolo Tonella.
\newblock Testing of deep reinforcement learning agents with surrogate models.
\newblock {\em {ACM} Trans. Softw. Eng. Methodol.}, 33(3):73:1--73:33, 2024.

\bibitem[\protect\citeauthoryear{Chandak \bgroup \em et al.\egroup }{2021}]{DBLP:conf/nips/ChandakNSLBT21}
Yash Chandak, Scott Niekum, Bruno~C. da~Silva, Erik~G. Learned{-}Miller, Emma Brunskill, and Philip~S. Thomas.
\newblock Universal off-policy evaluation.
\newblock In {\em NeurIPS}, pages 27475--27490, 2021.

\bibitem[\protect\citeauthoryear{Chaslot \bgroup \em et al.\egroup }{2008}]{DBLP:conf/aiide/ChaslotBSS08}
Guillaume Chaslot, Sander Bakkes, Istvan Szita, and Pieter Spronck.
\newblock Monte-carlo tree search: {A} new framework for game {AI}.
\newblock In {\em AIIDE}. The {AAAI} Press, 2008.

\bibitem[\protect\citeauthoryear{Chen \bgroup \em et al.\egroup }{2018}]{DBLP:journals/csur/ChenKLPTTZ18}
Tsong~Yueh Chen, Fei{-}Ching Kuo, Huai Liu, Pak{-}Lok Poon, Dave Towey, T.~H. Tse, and Zhi~Quan Zhou.
\newblock Metamorphic testing: {A} review of challenges and opportunities.
\newblock {\em {ACM} Comput. Surv.}, 51(1):4:1--4:27, 2018.

\bibitem[\protect\citeauthoryear{Cohen}{1995}]{DBLP:conf/icml/Cohen95}
William~W. Cohen.
\newblock Fast effective rule induction.
\newblock In {\em ICML}, pages 115--123. Morgan Kaufmann, 1995.

\bibitem[\protect\citeauthoryear{Delfosse \bgroup \em et al.\egroup }{2023}]{DBLP:conf/nips/DelfosseSDK23}
Quentin Delfosse, Hikaru Shindo, Devendra~Singh Dhami, and Kristian Kersting.
\newblock Interpretable and explainable logical policies via neurally guided symbolic abstraction.
\newblock In {\em NeurIPS}, 2023.

\bibitem[\protect\citeauthoryear{DeNero and Klein}{2010}]{berkeley_pacman}
John DeNero and Dan Klein.
\newblock Teaching introductory artificial intelligence with pac-man.
\newblock {\em Proceedings of the AAAI Conference on Artificial Intelligence}, 24(3):1885--1889, July 2010.

\bibitem[\protect\citeauthoryear{Dong \bgroup \em et al.\egroup }{2019}]{DBLP:conf/iclr/DongMLWLZ19}
Honghua Dong, Jiayuan Mao, Tian Lin, Chong Wang, Lihong Li, and Denny Zhou.
\newblock Neural logic machines.
\newblock In {\em {ICLR}}. OpenReview.net, 2019.

\bibitem[\protect\citeauthoryear{Dwarakanath \bgroup \em et al.\egroup }{2018}]{DBLP:conf/issta/DwarakanathASRB18}
Anurag Dwarakanath, Manish Ahuja, Samarth Sikand, Raghotham~M. Rao, R.~P. Jagadeesh~Chandra Bose, Neville Dubash, and Sanjay Podder.
\newblock Identifying implementation bugs in machine learning based image classifiers using metamorphic testing.
\newblock In {\em {ISSTA}}, pages 118--128. {ACM}, 2018.

\bibitem[\protect\citeauthoryear{Falcone}{2010}]{DBLP:conf/rv/Falcone10}
Yli{\`{e}}s Falcone.
\newblock You should better enforce than verify.
\newblock In {\em {RV}}, volume 6418 of {\em Lecture Notes in Computer Science}, pages 89--105. Springer, 2010.

\bibitem[\protect\citeauthoryear{Giacomo \bgroup \em et al.\egroup }{2020}]{DBLP:conf/aaai/GiacomoIFP20}
Giuseppe~De Giacomo, Luca Iocchi, Marco Favorito, and Fabio Patrizi.
\newblock Restraining bolts for reinforcement learning agents.
\newblock In {\em AAAI}, pages 13659--13662. {AAAI} Press, 2020.

\bibitem[\protect\citeauthoryear{Guo and Wei}{2022}]{9925786}
Wei Guo and Peng Wei.
\newblock Explainable deep reinforcement learning for aircraft separation assurance.
\newblock In {\em DASC}, pages 1--10, 2022.

\bibitem[\protect\citeauthoryear{Guo \bgroup \em et al.\egroup }{2020}]{DBLP:conf/kbse/GuoXLZLLS20}
Qianyu Guo, Xiaofei Xie, Yi~Li, Xiaoyu Zhang, Yang Liu, Xiaohong Li, and Chao Shen.
\newblock Audee: Automated testing for deep learning frameworks.
\newblock In {\em {ASE}}, pages 486--498. {IEEE}, 2020.

\bibitem[\protect\citeauthoryear{Jiang and Li}{2016}]{DBLP:conf/icml/JiangL16}
Nan Jiang and Lihong Li.
\newblock Doubly robust off-policy value evaluation for reinforcement learning.
\newblock In {\em {ICML}}, volume~48 of {\em {JMLR} Workshop and Conference Proceedings}, pages 652--661. JMLR.org, 2016.

\bibitem[\protect\citeauthoryear{Jiang and Luo}{2019}]{DBLP:conf/icml/JiangL19}
Zhengyao Jiang and Shan Luo.
\newblock Neural logic reinforcement learning.
\newblock In {\em {ICML}}, volume~97 of {\em Proceedings of Machine Learning Research}, pages 3110--3119. {PMLR}, 2019.

\bibitem[\protect\citeauthoryear{Leurent}{2018}]{highway_env}
Edouard Leurent.
\newblock An environment for autonomous driving decision-making.
\newblock \url{https://github.com/eleurent/highway-env}, 2018.

\bibitem[\protect\citeauthoryear{Li \bgroup \em et al.\egroup }{2023}]{DBLP:conf/kbse/LiWZCCZXMZ23}
Zhuo Li, Xiongfei Wu, Derui Zhu, Mingfei Cheng, Siyuan Chen, Fuyuan Zhang, Xiaofei Xie, Lei Ma, and Jianjun Zhao.
\newblock Generative model-based testing on decision-making policies.
\newblock In {\em {ASE}}, pages 243--254. {IEEE}, 2023.

\bibitem[\protect\citeauthoryear{Lundberg and Lee}{2017}]{DBLP:conf/nips/LundbergL17}
Scott~M. Lundberg and Su{-}In Lee.
\newblock A unified approach to interpreting model predictions.
\newblock In {\em NeurIPS}, pages 4765--4774, 2017.

\bibitem[\protect\citeauthoryear{{Marco Tulio Ribeiro et al.}}{2021}]{python_lime}
{Marco Tulio Ribeiro et al.}
\newblock Python implementation of {LIME}, 2021.
\newblock Available at GitHub: \url{https://github.com/marcotcr/lime}.

\bibitem[\protect\citeauthoryear{Milani \bgroup \em et al.\egroup }{2022}]{DBLP:conf/pkdd/MilaniZTSKPF22}
Stephanie Milani, Zhicheng Zhang, Nicholay Topin, Zheyuan~Ryan Shi, Charles~A. Kamhoua, Evangelos~E. Papalexakis, and Fei Fang.
\newblock {MAVIPER:} learning decision tree policies for interpretable multi-agent reinforcement learning.
\newblock In {\em {ECML} {PKDD}}, volume 13716 of {\em Lecture Notes in Computer Science}, pages 251--266. Springer, 2022.

\bibitem[\protect\citeauthoryear{Milani \bgroup \em et al.\egroup }{2024}]{DBLP:journals/csur/MilaniTVF24}
Stephanie Milani, Nicholay Topin, Manuela Veloso, and Fei Fang.
\newblock Explainable reinforcement learning: {A} survey and comparative review.
\newblock {\em {ACM} Comput. Surv.}, 56(7):168:1--168:36, 2024.

\bibitem[\protect\citeauthoryear{Mnih \bgroup \em et al.\egroup }{2015}]{DBLP:journals/nature/MnihKSRVBGRFOPB15}
Volodymyr Mnih, Koray Kavukcuoglu, David Silver, Andrei~A. Rusu, Joel Veness, Marc~G. Bellemare, Alex Graves, Martin~A. Riedmiller, Andreas Fidjeland, Georg Ostrovski, Stig Petersen, Charles Beattie, Amir Sadik, Ioannis Antonoglou, Helen King, Dharshan Kumaran, Daan Wierstra, Shane Legg, and Demis Hassabis.
\newblock Human-level control through deep reinforcement learning.
\newblock {\em Nat.}, 518(7540):529--533, 2015.

\bibitem[\protect\citeauthoryear{Molnar}{2022}]{molnar2022}
Christoph Molnar.
\newblock {\em Interpretable Machine Learning}.
\newblock 2 edition, 2022.

\bibitem[\protect\citeauthoryear{Nielsen}{2019}]{alpha_zero_news}
Peter~Heine Nielsen.
\newblock The exciting impact of a game changer: When magnus met alphazero, 2019.
\newblock Appeared in {New in Chess} and published online at \url{https://www.newinchess.com/media/wysiwyg/product_pdf/872.pdf}.

\bibitem[\protect\citeauthoryear{Odriozola{-}Olalde \bgroup \em et al.\egroup }{2023}]{DBLP:conf/sii/OdriozolaOlaldeZA23}
Haritz Odriozola{-}Olalde, Maider Zamalloa, and Nestor Arana{-}Arexolaleiba.
\newblock Shielded reinforcement learning: {A} review of reactive methods for safe learning.
\newblock In {\em {SII}}, pages 1--8. {IEEE}, 2023.

\bibitem[\protect\citeauthoryear{{Oriol Vinyals et al.}}{2019}]{DBLP:journals/nature/VinyalsBCMDCCPE19}
{Oriol Vinyals et al.}
\newblock Grandmaster level in starcraft {II} using multi-agent reinforcement learning.
\newblock {\em Nat.}, 575(7782):350--354, 2019.

\bibitem[\protect\citeauthoryear{Raffin \bgroup \em et al.\egroup }{2021}]{sb3}
Antonin Raffin, Ashley Hill, Adam Gleave, Anssi Kanervisto, Maximilian Ernestus, and Noah Dormann.
\newblock Stable-baselines3: Reliable reinforcement learning implementations.
\newblock {\em Journal of Machine Learning Research}, 22(268):1--8, 2021.

\bibitem[\protect\citeauthoryear{Ribeiro \bgroup \em et al.\egroup }{2016}]{DBLP:conf/kdd/Ribeiro0G16}
Marco~T{\'{u}}lio Ribeiro, Sameer Singh, and Carlos Guestrin.
\newblock "why should {I} trust you?": Explaining the predictions of any classifier.
\newblock In {\em SIGKDD}, pages 1135--1144. {ACM}, 2016.

\bibitem[\protect\citeauthoryear{Schulman \bgroup \em et al.\egroup }{2017}]{DBLP:journals/corr/SchulmanWDRK17}
John Schulman, Filip Wolski, Prafulla Dhariwal, Alec Radford, and Oleg Klimov.
\newblock Proximal policy optimization algorithms.
\newblock {\em CoRR}, abs/1707.06347, 2017.

\bibitem[\protect\citeauthoryear{Silver \bgroup \em et al.\egroup }{2016}]{DBLP:journals/nature/SilverHMGSDSAPL16}
David Silver, Aja Huang, Chris~J. Maddison, Arthur Guez, Laurent Sifre, George van~den Driessche, Julian Schrittwieser, Ioannis Antonoglou, Vedavyas Panneershelvam, Marc Lanctot, Sander Dieleman, Dominik Grewe, John Nham, Nal Kalchbrenner, Ilya Sutskever, Timothy~P. Lillicrap, Madeleine Leach, Koray Kavukcuoglu, Thore Graepel, and Demis Hassabis.
\newblock Mastering the game of go with deep neural networks and tree search.
\newblock {\em Nat.}, 529(7587):484--489, 2016.

\bibitem[\protect\citeauthoryear{Silver \bgroup \em et al.\egroup }{2018}]{doi:10.1126/science.aar6404}
David Silver, Thomas Hubert, Julian Schrittwieser, Ioannis Antonoglou, Matthew Lai, Arthur Guez, Marc Lanctot, Laurent Sifre, Dharshan Kumaran, Thore Graepel, Timothy Lillicrap, Karen Simonyan, and Demis Hassabis.
\newblock A general reinforcement learning algorithm that masters chess, shogi, and go through self-play.
\newblock {\em Science}, 362(6419):1140--1144, 2018.

\bibitem[\protect\citeauthoryear{Tappler \bgroup \em et al.\egroup }{2022}]{DBLP:conf/ijcai/TapplerCAK22}
Martin Tappler, Filip~Cano C{\'{o}}rdoba, Bernhard~K. Aichernig, and Bettina K{\"{o}}nighofer.
\newblock Search-based testing of reinforcement learning.
\newblock In {\em {IJCAI}}, pages 503--510. ijcai.org, 2022.

\bibitem[\protect\citeauthoryear{Tappler \bgroup \em et al.\egroup }{2024}]{DBLP:conf/icst/TapplerMAK24}
Martin Tappler, Edi Muskardin, Bernhard~K. Aichernig, and Bettina K{\"{o}}nighofer.
\newblock Learning environment models with continuous stochastic dynamics - with an application to deep {RL} testing.
\newblock In {\em {IEEE} Conference on Software Testing, Verification and Validation, {ICST} 2024, Toronto, ON, Canada, May 27-31, 2024}, pages 197--208. {IEEE}, 2024.

\bibitem[\protect\citeauthoryear{Towers \bgroup \em et al.\egroup }{2024}]{towers2024gymnasiumstandardinterfacereinforcement}
Mark Towers, Ariel Kwiatkowski, Jordan Terry, John~U. Balis, Gianluca~De Cola, Tristan Deleu, Manuel Goulão, Andreas Kallinteris, Markus Krimmel, Arjun KG, Rodrigo Perez-Vicente, Andrea Pierré, Sander Schulhoff, Jun~Jet Tai, Hannah Tan, and Omar~G. Younis.
\newblock Gymnasium: A standard interface for reinforcement learning environments, 2024.

\bibitem[\protect\citeauthoryear{Uehara \bgroup \em et al.\egroup }{2022}]{DBLP:journals/corr/abs-2212-06355}
Masatoshi Uehara, Chengchun Shi, and Nathan Kallus.
\newblock A review of off-policy evaluation in reinforcement learning.
\newblock {\em CoRR}, abs/2212.06355, 2022.

\bibitem[\protect\citeauthoryear{van Hasselt \bgroup \em et al.\egroup }{2016}]{DBLP:conf/aaai/HasseltGS16}
Hado van Hasselt, Arthur Guez, and David Silver.
\newblock Deep reinforcement learning with double q-learning.
\newblock In {\em AAAI}, pages 2094--2100. {AAAI} Press, 2016.

\bibitem[\protect\citeauthoryear{Watkins and Dayan}{1992}]{DBLP:journals/ml/WatkinsD92}
Christopher J. C.~H. Watkins and Peter Dayan.
\newblock Technical note q-learning.
\newblock {\em Mach. Learn.}, 8:279--292, 1992.

\bibitem[\protect\citeauthoryear{Welch}{1947}]{welch}
B.~L. Welch.
\newblock The generalization of `student's' problem when several different population variances are involved.
\newblock {\em Biometrika}, 34(1/2):28--35, 1947.

\bibitem[\protect\citeauthoryear{Xie \bgroup \em et al.\egroup }{2011}]{DBLP:journals/jss/XieHMKXC11}
Xiaoyuan Xie, Joshua Wing~Kei Ho, Christian Murphy, Gail~E. Kaiser, Baowen Xu, and Tsong~Yueh Chen.
\newblock Testing and validating machine learning classifiers by metamorphic testing.
\newblock {\em J. Syst. Softw.}, 84(4):544--558, 2011.

\bibitem[\protect\citeauthoryear{Zambaldi \bgroup \em et al.\egroup }{2018}]{DBLP:journals/corr/abs-1806-01830}
Vin{\'{\i}}cius~Flores Zambaldi, David Raposo, Adam Santoro, Victor Bapst, Yujia Li, Igor Babuschkin, Karl Tuyls, David~P. Reichert, Timothy~P. Lillicrap, Edward Lockhart, Murray Shanahan, Victoria Langston, Razvan Pascanu, Matthew~M. Botvinick, Oriol Vinyals, and Peter~W. Battaglia.
\newblock Relational deep reinforcement learning.
\newblock {\em CoRR}, abs/1806.01830, 2018.

\bibitem[\protect\citeauthoryear{Zhang \bgroup \em et al.\egroup }{2022}]{DBLP:journals/tse/ZhangHML22}
Jie~M. Zhang, Mark Harman, Lei Ma, and Yang Liu.
\newblock Machine learning testing: Survey, landscapes and horizons.
\newblock {\em {IEEE} Trans. Software Eng.}, 48(2):1--36, 2022.

\bibitem[\protect\citeauthoryear{Zolfagharian \bgroup \em et al.\egroup }{2023}]{DBLP:journals/tse/ZolfagharianABBS23}
Amirhossein Zolfagharian, Manel Abdellatif, Lionel~C. Briand, Mojtaba Bagherzadeh, and Ramesh S.
\newblock A search-based testing approach for deep reinforcement learning agents.
\newblock {\em {IEEE} Trans. Software Eng.}, 49(7):3715--3735, 2023.

\end{thebibliography}
\newpage
\appendix{
\section{Metamorphic Relation -- Extended Example}

We provide an extended example of metamorphic relations to give more insights into our MT approach. 
This extended example provides additional information on the small PAC-Man environment's features, actions, and metamorphic relations.

We have five actions available to the agent represented by integers:
\begin{itemize}
    \item $0$ represents going \emph{North}
    \item $1$ represents going \emph{South}
    \item $2$ represents going \emph{East}
    \item $3$ represents going \emph{West} 
    \item $4$ represents \emph{stopping}
\end{itemize}
We use metamorphic relations to express 90-degree clockwise rotation, where we showed a portion of the feature relations in the main part of the paper. To ``rotate all features'', we use the same feature relations for every pair $(a_1,a_2)$ of actions in $AP = \{(0,2), (2,1), (1,3), (3,0)\}$. 
Each set feature relations $MR(a_1,a_2)$ is a metamorphic relation given by:
\begin{align*}
MR(a_1,a_2) = &\big \{ \\
\intertext{The features $2$ to $5$ represent the one-hot encoding of the direction toward the closest capsule:}
&\langle a_1,a_2, 2,4, = \rangle &\langle a_1,a_2, 3,5, = \rangle\\
&\langle a_1,a_2, 4,3, = \rangle &\langle a_1,a_2, 5,2, = \rangle\\
\intertext{The features $9$ to $12$ represent the one-hot encoding of the direction toward the closest food:}
&\langle a_1,a_2, 9,11, = \rangle &\langle a_1,a_2, 10,12, = \rangle\\
&\langle a_1,a_2, 11,10, = \rangle &\langle a_1,a_2, 12,9, = \rangle\\
\intertext{The features $15$ to $22$ represent the one-hot encoding of the angle to Ghost 0 with a resolution of 8:}
&\langle a_1,a_2, 15,17, = \rangle &\langle a_1,a_2, 16,18, = \rangle\\
&\langle a_1,a_2, 17,19, = \rangle &\langle a_1,a_2, 18,20, = \rangle\\
&\langle a_1,a_2, 19,21, = \rangle &\langle a_1,a_2, 20,22, = \rangle\\
&\langle a_1,a_2, 21,15, = \rangle &\langle a_1,a_2, 22,16, = \rangle\\
\intertext{The features $23$ to $26$ represent the one-hot encoding of the direction toward Ghost 0:}
&\langle a_1,a_2, 23,25, = \rangle &\langle a_1,a_2, 24,26, = \rangle\\
&\langle a_1,a_2, 25,24, = \rangle &\langle a_1,a_2, 26,23, = \rangle\\
\intertext{The features $29$ to $32$ represent the one-hot encoding of the direction toward which Ghost 0 is heading:}
&\langle a_1,a_2, 29,31, = \rangle &\langle a_1,a_2, 30,32, = \rangle\\
&\langle a_1,a_2, 31,30, = \rangle &\langle a_1,a_2, 32,29, = \rangle\\
\intertext{The features $39$ to $46$ represent the one-hot encoding of the angle to Ghost 1 with a resolution of 8:}
&\langle a_1,a_2, 39,41, = \rangle &\langle a_1,a_2, 40,42, = \rangle\\
&\langle a_1,a_2, 41,43, = \rangle &\langle a_1,a_2, 42,44, = \rangle\\
&\langle a_1,a_2, 43,45, = \rangle &\langle a_1,a_2, 44,46, = \rangle\\
&\langle a_1,a_2, 45,39, = \rangle &\langle a_1,a_2, 46,40, = \rangle\\
\intertext{The features $47$ to $50$ represent the one-hot encoding of the direction toward Ghost 1:}
&\langle a_1,a_2, 47,49, = \rangle &\langle a_1,a_2, 48,50, = \rangle\\
&\langle a_1,a_2, 49,48, = \rangle &\langle a_1,a_2, 50,47, = \rangle\\
\intertext{The features $53$ to $56$ represent the one-hot encoding of the direction toward which Ghost 1 is heading:}
&\langle a_1,a_2, 53,55, = \rangle &\langle a_1,a_2, 54,56, = \rangle\\
&\langle a_1,a_2, 55,54, = \rangle &\langle a_1,a_2, 56,53, = \rangle\\
\intertext{The features $62$ to $65$ are Boolean flags that denote if moving into one of the four cardinal directions is possible:}
&\langle a_1,a_2, 62,64, = \rangle &\langle a_1,a_2, 63,65, = \rangle\\
&\langle a_1,a_2, 64,63, = \rangle &\langle a_1,a_2, 65,62, = \rangle\\
\intertext{The features $C = \{67,68,36,37,60,61\}$ specify the x,y-coordinates of the PAC-Man, Ghost 0, and Ghost 1. Since they carry little information compared to the other features above, we want to ignore the coordinates. This can be expressed as follows for $c \in C$:}
&\langle a_1,a_2, c,c, \mathbb{R}  \times \mathbb{R}\rangle \\
&\big\}
\end{align*}

The features representing a direction with a resolution of $4$ basically specify the action to take the agent closer to the respective object, while respecting walls. Angle features have a resolution of 8 and do not respect walls, for example, $f_{15}^s = 1$ specifies that Ghost $0$ is in the north of the agent, while there may be a wall directly to the north. The final six feature relations describe that every coordinate value is related to every coordinate value, thus essentially ignoring the values. To avoid creating excessively many rules and improve, we implement this as a special case when generalizing rules through $M_{gen}$ with $M = \{MR(a_1,a_2) \mid (a_1,a_2) \in AP\}$, where we remove conditions on features with indexes in $C$. 

\section{Hyperparameters}
We trained RL policies using stable-baselines3~\cite{sb3}. For the PAC-Man environments, we used the following hyperparameters for DQN policies and left the rest at their default values:
\begin{itemize}
    \item \verb+batch_size=256+
    \item \verb+buffer_size=50_000+,
    \item \verb+exploration_fraction=0.5+
    \item \verb+gamma = 0.95+
    \item \verb+gradient_steps=-1+
    \item \verb+policy="MlpPolicy"+ with two layers of fully connected hidden layers with $256$ neuron and the \texttt{ReLU} activation function
\end{itemize}
To train PPO policies for the example below, we used the following hyperparameters, also leaving the rest at their default values:
\begin{itemize}
    \item \verb+vf_coef= 0.75+
    \item \verb+ent_coef= 0.0+
    \item \verb+clip_range=0.5+
    \item \verb+learning_rate=2.5e-4+
    \item \verb+batch_size=256+
    \item \verb+gamma=0.9+
    \item \verb+n_epochs=10+
    \item \verb+n_steps=256+
    \item \verb+policy="MlpPolicy"+ with two layers of fully connected hidden layers with $256$ neuron and the \texttt{ReLU} activation function for both the policy and the value network
\end{itemize}

For the \texttt{highway-env} environments, we only trained DQN policies with the following hyperparameter changes:
\begin{itemize}
     \item \verb+learning_rate=5e-4+
     \item \verb+buffer_size=15000+
     \item \verb+learning_starts=200+
     \item \verb+batch_size=32+
     \item \verb+gamma=0.8+
     \item \verb+train_freq=1+
     \item \verb+gradient_steps=1+
     \item \verb+target_update_interval=1000+
         \item \verb+policy="MlpPolicy"+ with two layers of fully connected hidden layers with $256$ neurons and the \texttt{ReLU} activation function for both the policy and the value network
\end{itemize}

\section{Explaining Policies}

\begin{table}[t]  
\footnotesize
\centering
        \begin{tabular}{l|l|l|l|l}
        \multirow{2}{*}{Experiment} & \multicolumn{2}{|c|}{DQN} & \multicolumn{2}{|c}{PPO} \\ \cline{2-5}
         &  Ghosts & Scared &  Ghosts & Scared \\ \hline 
         small & 0.82 & 0.60 &  0.78 & 0.21  \\ \hline 
medium & 0.72 & 0.45 &  0.67 & 0.24  \\ \hline 
original & 0.62 & 0.13 & 0.50 & 0.05  \\  

    \end{tabular}
    \caption{Ratio of specific features in rules mined from DQN-trained and PPO-trained policies in three PAC-Man levels.
    The DQN-trained policies recognize the relevance of ghosts being scared more often than PPO-trained policies}
    \label{tab:x-reckless}
\end{table}

In the following, we provide an example of explaining the decision-making of policies with mined rules. In the example, we compare the decision-making of PPO~\cite{DBLP:journals/corr/SchulmanWDRK17} and DQN policies, where we trained PPO policies for the same number of steps. 
For mining negative rules from PPO policies, the inclusion sets for action $a$ comprise those experiences where $a$ has the minimal probability of being chosen rather than the minimal Q-value as proposed in Sect.~\ref{sec:mine_rules} for Q-learning-based policies. Note that while rules can be mined, rule-guided execution does not work well for on-policy algorithms, like PPO. The complete hyperparameter configuration can be found in the appendix.

In manual analyses of executions, we found that PPO-trained policies behave "recklessly" in the PAC-Man environments. In general, both DQN and PPO policies steer the agents to capsules, which makes the ghosts \emph{scared} and vulnerable for a fixed amount of time. Since killing ghosts grants a large positive reward, the policies learn to chase ghosts. Compared to DQN, we found that PPO does so very aggressively such that the agent often collides with non-scared ghosts resulting in an unsuccessful episode termination. 

The mined rules give insights into why this happens, as we found that rules mined from PPO policies rarely place constraints on whether a ghost is scared.
Table~\ref{tab:x-reckless} shows exact ratios of how often ghosts-, and ghost-scared-related features appear in rules. It can be seen that many rule conditions center around features related to ghosts for DQN and PPO alike.
When we look at specific features describing if and how long a specific ghost is scared we consistently see large differences between DQN and PPO. Hence, this can explain why PPO policies do not care if they collide with an invincible ghost, as those policies consider ghosts as often as DQN, but rarely consider if they are scared.

\section{Code and Data}
Our implementation of \algname{} and the presented experimental data are available at \url{https://doi.org/10.6084/m9.figshare.28569017}. 
% This is currently a temporary private link to ensure anonymity. We use an external website because the experimental data, even in compressed form, requires about $400$ MB of space, which exceeds the upload size limit of CMT. 
The experimental data can be found in the directory \texttt{reference\_pickles/eval\_stats}, stored using Python's pickle module. The directory \texttt{reference\_pickles} further contains trained models. The directories \texttt{run\_scripts\_hw} and \texttt{run\_scripts\_pac} contain bash scripts, which call the functionality of our implementation using the same parameter configuration as in the reported experiments. Please see the readme file for more information. 
}

\end{document}